\pdfoutput=1

\documentclass[11pt]{article}

\usepackage[preprint ]{acl}

\usepackage{times}
\usepackage{latexsym}
\usepackage{float}

\usepackage[T1]{fontenc}

\usepackage[utf8]{inputenc}

\usepackage{microtype}

\usepackage{inconsolata}

\usepackage{graphicx}

\title{MM-Eval: A Hierarchical Benchmark for\\ Modern Mongolian Evaluation in LLMs}

\author{
  \textbf{Mengyuan Zhang\textsuperscript{1,3,4,5}
  \thanks{\texttt{This work was done during the internship of Mengyuan Zhang at Yucang Technology.  \href{mailto:joenahm@yeah.net}{joenahm@yeah.net}}}}, 
 \textbf{Ruihui Wang\textsuperscript{2,4}},
 \textbf{Bo Xia\textsuperscript{3}},
 \textbf{Yuan Sun\textsuperscript{2,4,5}},
 \textbf{Xiaobing Zhao\textsuperscript{2,4,5}}
 \thanks{\texttt{{Xiaobing Zhao is the corresponding author. \href{mailto:nmzxb_cn@163.com}{nmzxb\_cn@163.com}}}}, 
\\
\\
 \textsuperscript
 {1}School of Philosophy and Religious Studies, Minzu University of China
 \\
 \textsuperscript
 {2}School of Information Engineering, Minzu University of China
 \\
 \textsuperscript
 {3}Yucang Technology, Beijing 100085, China
 \\
 \textsuperscript
 {4}National Language Resources Monitoring and Research Center for Ethnic Languages,\\ Minzu University of China
 \\
 \textsuperscript
 {5}Information Security Research Center Institute of National Security MUC,\\ Minzu University of China
}

\begin{document}
\maketitle
\begin{abstract}
Large language models (LLMs) excel in high-resource languages but face notable challenges in low-resource languages like Mongolian. This paper addresses these challenges by categorizing capabilities into language abilities (syntax and semantics) and cognitive abilities (knowledge and reasoning). To systematically evaluate these areas, we developed MM-Eval, a specialized dataset based on Modern Mongolian Language Textbook I and enriched with WebQSP and MGSM datasets.

Preliminary experiments on models including Qwen2-7B-Instruct, GLM4-9b-chat, Llama3.1-8B-Instruct, GPT-4, and DeepseekV2.5 revealed that: 1) all models performed better on syntactic tasks than semantic tasks, highlighting a gap in deeper language understanding; and 2) knowledge tasks showed a moderate decline, suggesting that models can transfer general knowledge from high-resource to low-resource contexts.

The release of MM-Eval—comprising 569 syntax, 677 semantics, 344 knowledge, and 250 reasoning tasks—offers valuable insights for advancing NLP and LLMs in low-resource languages like Mongolian. The dataset is available at \url{https://github.com/joenahm/MM-Eval}.
\end{abstract}

\section{Introduction}

\begin{figure*}[t]
  \caption {\label{figure1}Workflow for Constructing the MM-Eval Dataset}
    \centering 
    \includegraphics[width=1.0\textwidth]{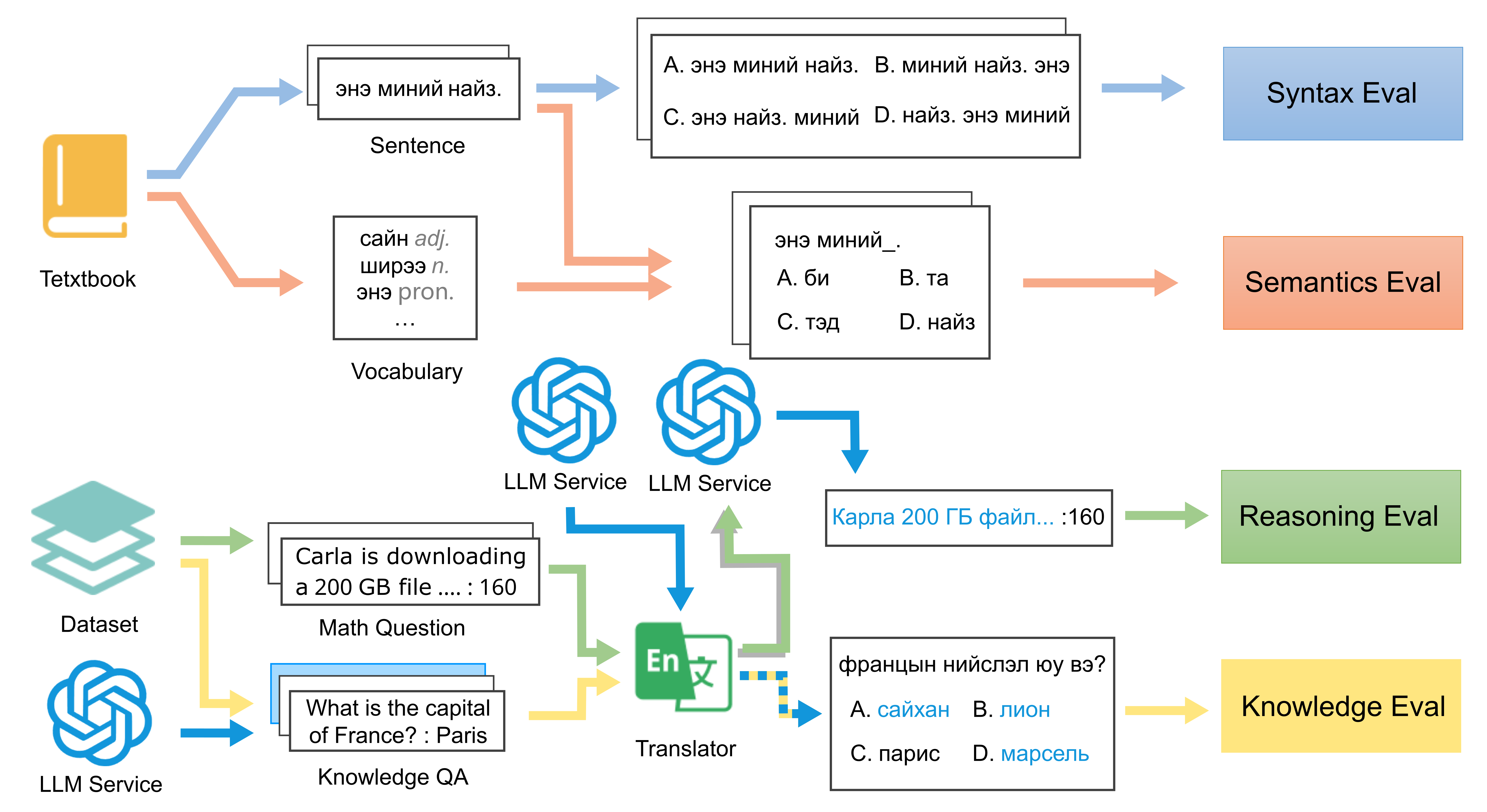}
    \renewcommand{\figurename}{Figure}
    \label{firstpic} 
\end{figure*}

In recent years, large language models (LLMs) have revolutionized natural language processing (NLP), demonstrating remarkable capabilities in understanding and generating human language, excelling in tasks such as context comprehension\cite{context}, language generation\cite{qg}, summarization\cite{summarization}, question answering\cite{QA}, and translation\cite{icml}. Models like ChatGPT\cite{gpt4} and Llama\cite{llama2} have set new benchmarks across a wide range of languages, primarily high-resource ones such as Chinese and English. However, the support for low-resource languages like Mongolian remains largely unexplored.

Mongolian, spoken by millions across Mongolia and Inner Mongolia of China, presents unique linguistic challenges due to its complex grammar, script, and historical evolution. In Mongolia, modern Mongolian is written using the Cyrillic script, based on the Russian alphabet, while in Inner Mongolia, China, the traditional Mongolian script, derived from the Sogdian-Uyghur script, is used. This paper focuses on modern Mongolian written in the Cyrillic script. Despite some efforts to include Mongolian in NLP research, there is still a significant gap in understanding how well LLMs can handle Mongolian across various linguistic dimensions.

This research aims to fill the gap in Mongolian language support by systematically evaluating modern LLMs' capabilities in processing Mongolian. Unlike existing task-oriented datasets, this study focuses on models proven effective in high-resource languages. For Mongolian, we adopt a linguistic perspective, constructing a dataset based on language proficiency levels and previous LLM performance. Our dataset is organized into four hierarchical levels: syntax, semantics, knowledge, and reasoning. This structure allows for a detailed evaluation of model performance at different proficiency levels, providing deeper insights into their strengths and limitations.

By uncovering both the strengths and weaknesses of current models, we aim to provide a benchmark for future Mongolian NLP research, contribute to the broader understanding of LLM support for low-resource languages, and help enhance their Mongolian language capabilities. We summarize the main lessons learned and our main contributions as follows:

\begin{itemize}
\item This paper introduces MM-Eval, a specialized dataset for evaluating the capabilities of large language models (LLMs) in modern Mongolian, which is a low-resource language. 
\item This paper proposes a Dual Capability Framework that evaluates LLMs by dividing their capabilities into language abilities (syntax and semantics) and cognitive abilities (knowledge and reasoning). This framework allows for a detailed understanding of model performance at different language proficiency levels.
\item This paper provides a comprehensive evaluation of LLMs in Mongolian, covering syntax, semantics, knowledge, and reasoning. This evaluation reveals the strengths and weaknesses of current models in processing Mongolian.
\end{itemize}

\section{Related Work}
As large-scale models continued to evolve, more comprehensive and diverse open test datasets, such as CValues\cite{CValues}, were introduced. These datasets covered specialized knowledge (MMLU\cite{MMLU}), logical reasoning (GPQA\cite{GPQA}), mathematical ability (GSM8K\cite{GSM8K}), coding skills (HumanEval\cite{HumanEval}), and instruction-following capabilities (LiveBench\cite{LiveBench}). These evaluation datasets can effectively test the model's various abilities in high-resource languages.

In contrast, Mongolian language processing has historically focused on downstream tasks such as text classification\cite{cino} and named entity recognition\cite{named}, with limited evaluation for large models. This work significantly fills that gap by introducing comprehensive evaluations tailored for Mongolian language models, addressing the deficiencies observed in previous studies.

\section{MM-Eval}

MM-Eval consists of four components: Syntax, Semantics, Knowledge, and Reasoning. The Syntax section contains 569 multiple-choice questions, the Semantics section includes 677 multiple-choice questions, the Knowledge section comprises 344 multiple-choice questions, and the Reasoning section features 250 math problems that require numerical answers. Figure \ref{figure1} illustrates the overall construction process of the MM-Eval dataset. For better image layout, the order of the knowledge evaluation and reasoning evaluation tasks has been swapped.

\subsection{Dual Capability Framework}
We developed a dataset with a Dual Capability Framework to evaluate LLMs by dividing their capabilities into language abilities and cognitive abilities. The cognitive abilities of a model are reflected through its primary training language, while language abilities vary depending on the language in question. To address this, our dataset is structured with a focus on these dual capabilities. Specifically, within language abilities, we distinguish between syntax and semantics. In terms of cognitive abilities, we further differentiate between knowledge and reasoning. 

The language abilities section of our dataset evaluates the model’s proficiency in Mongolian, a low-resource language, reflecting its mastery of Mongolian independent of other language training data. In contrast, the cognitive abilities section assesses the model’s overall cognitive capacity, which is influenced by all its training data but applied to Mongolian. This section highlights the alignment between Mongolian and the model's primary training language, showcasing how cognitive capabilities are transferred and manifested in Mongolian.

\subsection{Data Collection}
The primary source of our data for the language abilities section is \textit{Modern Mongolian Language Textbook I} \cite{MMLT1}. We selected sentences from the dialogues and texts within this book to construct datasets for both syntax and semantics parts of evaluation. For the cognitive abilities section, the knowledge data is derived from two sources: a portion comes from the WebQSP\cite{WebQSP} dataset, which includes information related to geography and country-specific knowledge, and the other portion is generated using heuristic rules with ChatGPT API, followed by manual proofreading for accuracy. The reasoning data is sourced from the MGSM\cite{MGSM} dataset, focusing on cognitive reasoning tasks.

\subsection{Data Processing}
The content from the textbook is initially obtained through OCR to create an electronic text version. Subsequently, we extract dialogues, texts, and vocabulary from each lesson and perform data cleaning and manual correction to ensure accuracy. During this process, sentences that are excessively long, too short, or irrelevant for evaluation purposes are removed. For the WebQSP and MGSM datasets, a straightforward JSON format conversion is applied, with each question mapped to a corresponding answer.

\subsection{Syntax Eval}
For the syntax evaluation dataset, we selected sentences with three or more words from the extracted dialogue content in the textbook. After deduplication, the sentences were split by spaces, and their word order was shuffled to generate three incorrect options. These options were manually verified to ensure syntactic errors. Each original sentence, along with the three syntactically incorrect options, formed a multiple-choice question with four options.

\subsection{Semantics Eval}
For the semantic evaluation dataset, we utilized sentences from both the dialogues and texts in the textbook, as well as the vocabulary lists. Each vocabulary list provides key terms for each lesson, along with part-of-speech information, making it well-suited for constructing semantic knowledge questions. Given the rich morphological variation in Mongolian, particularly with verbs, we limited our selection to nouns, pronouns, adjectives, and adverbs to avoid potential issues stemming from complex inflectional forms that could compromise the accuracy of the questions themselves.

First, each sentence was split by spaces, and the resulting tokens were matched against the vocabulary list. Sentences without any matched words were discarded. For each sentence with a match, one of the matched words was randomly selected as the correct answer and removed from the sentence. Based on the part of speech of the removed word, three distractor words were selected from the vocabulary list. Specifically, nouns and pronouns were used as distractors for each other, while adjectives and adverbs were used similarly. The key criterion was that the distractor words should be plausible yet definitively incorrect as the answer. This process resulted in the construction of semantic evaluation questions.

\begin{figure*}
  \caption {\label{figure2}
  Performance of various models across four evaluation dimensions}

    \centering 
    \includegraphics[width=1.0\textwidth]{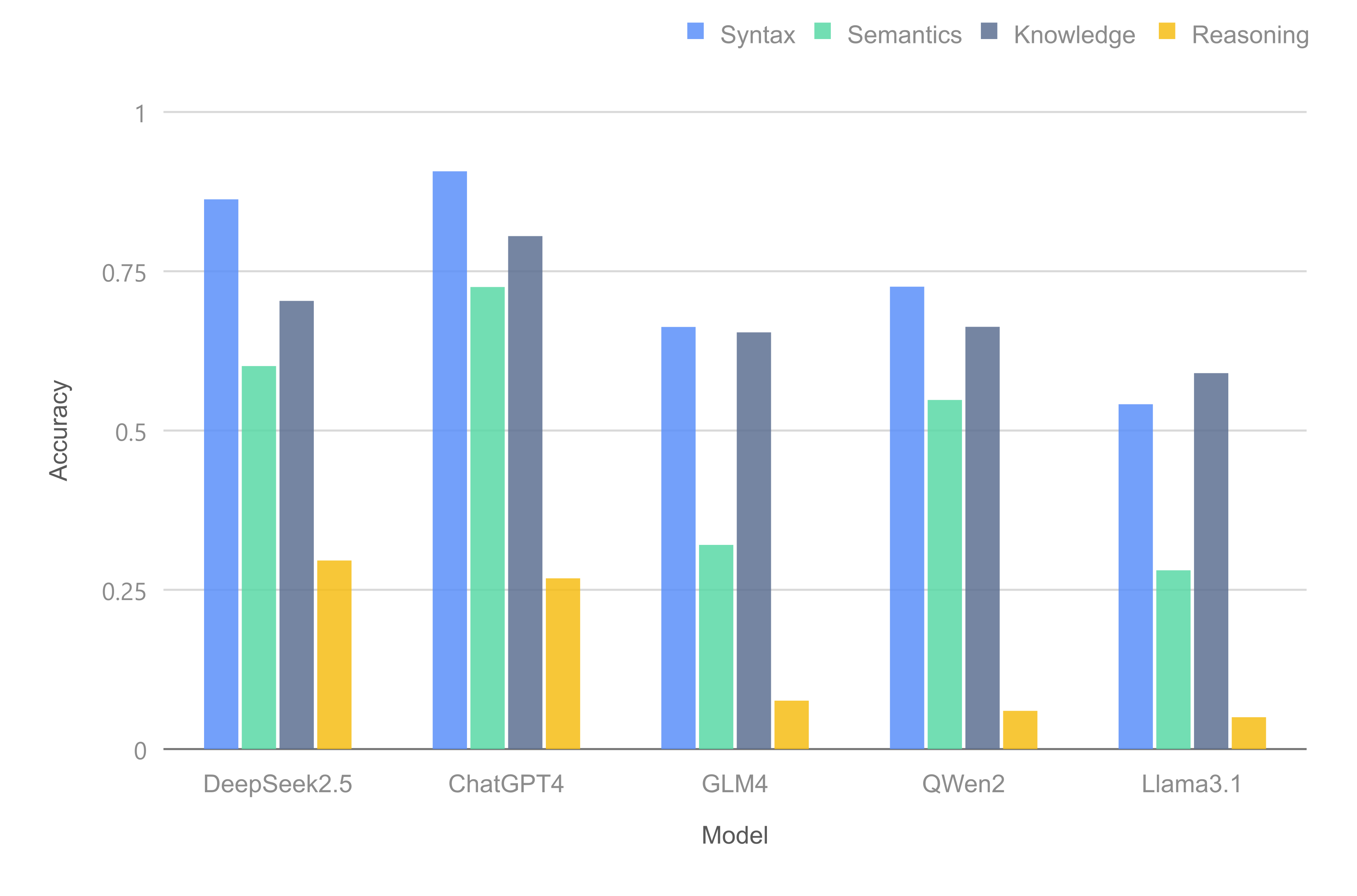}
    \renewcommand{\figurename}{Figure}
    \label{firstpic} 
\end{figure*}

\begin{table*}[t]
  \centering
  \begin{tabular}{lllll}
    \hline
    \textbf{Model}           & \textbf{Syntax} & \textbf{Semantic} & \textbf{Knowledge} & \textbf{Reasoning} \\
    \hline
    deepseekv2.5    & 86.29\%    & 60.12\%    & 70.35\%  &\textbf{29.6}\%    \\
chatgpt4-turbo      & \textbf{90.69\%}   & \textbf{72.53\%}   & \textbf{80.52\%}        & 26.8\%                                 \\
glm4-9b-chat          & 66.26\%  & 32.05\%         & 65.41\%   & 7.6\%            \\
qwen2-7b-instruct     & 72.58\%    & 54.8\%       & 66.28\%   & 6\%                \\
llama-3.1-8b-instruct & 54.13\%    & 28.06\%   & 59.01\%      & 5\%        \\ 
    \hline
  \end{tabular}
  \caption{\label{table1}
   Model accuracy comparison across four MM-Eval evaluation dimensions
  }
\end{table*}

\subsection{Knowledge Eval}
For the Knowledge Evaluation dataset, one of our data sources is the WebQSP dataset. However, due to the complexity of the knowledge types contained within it, we focused exclusively on portions that qualify as common knowledge. Specifically, we filtered the data using the "InferentialChain" field. This filtering yielded common knowledge related to countries, including their continents, capitals, official languages, currencies, and the flow directions of notable rivers.

Another data source was generated using heuristic methods leveraging ChatGPT API to create data on topics such as dates, chemical element symbols, simple arithmetic operations, and general life knowledge.
These entries were formatted as question-answer pairs and expressed in English. After merging these datasets, we utilized the ChatGPT API to generate three similar incorrect answers based on the provided responses. Each question's answer underwent manual verification, and any problematic incorrect options were adjusted accordingly. Subsequently, we employed the translation API from NiuTrans.com to translate the English content into Mongolian, ensuring that the translated results were also subject to manual quality checks.

\subsection{Reasoning Eval}
For the Reasoning Evaluation dataset, the sources are the English and Chinese versions of the MGSM dataset, which contain identical content and answers. This dataset comprises application-style mathematical problems, each accompanied by numeric answers. We adopted a translation approach to convert both the English and Chinese versions of the questions into Mongolian. Given the complex logical relationships inherent in these mathematical problems, ensuring the accuracy of the translated content is paramount.

Our strategy involved submitting the original English and Chinese texts alongside their respective Mongolian translations to the ChatGPT API for comparative evaluation. This process allowed us to assess the accuracy of the translations. In cases where the translations were found to be inaccurate or inadequately expressed, we made modifications based on the original texts, resulting in accurate Mongolian translations. Finally, we conducted a manual quality check to ensure the overall accuracy and clarity of the translated content.

\section{Experiments}
\subsection{Setup}
We deployed and inferred a local open-source model on a NVIDIA Tesla V100 (32GB) device. The inference parameters are: temperature=0, top-p=0.1, frequency penalty=1. The closed-source models are all invoked using APIs. The system prompt used for inference is: "\textit{You are an AI assistant proficient in Mongolian.}". There are different user prompts for four different tasks, namely: Syntax:\textit{"Select the grammatically correct sentence that follows the rules of Mongolian expression from the options below, and return the corresponding letter of the option (such as A, B, C, or D), do not return anything else."}; Semantic:\textit{"Complete the sentence to make it grammatically correct and meaningful in Mongolian. Return only the letter of the correct option (A, B, C, or D), do not return anything else."}; Knowledge:\textit{"Based on the following question, choose the correct answer.Return only the letter of the correct option (A, B, C, or D), do not return anything else."}; Reasoning:\textit{"Calculate the result: Perform the calculations based on the given mathematical problem."}

\subsection{Models}
We selected current mainstream open-source and closed-source models as the test models for our experimental dataset. The open-source models are: Qwen2-7B-Instruct\cite{qwen}, GLM4-9b-chat\cite{glm}, Llama3.1-8B-Instruct. The closed-source models are: GPT-4-Turbo-04-09, DeepseekV2.5\cite{deepseek}. The input data is the question from the dataset. The evaluation metric is Accuracy.

\subsection{Results}

Table \ref{table1} presents the corresponding results of different models in four evaluation directions. Figure \ref{figure2} shows the specific performance of different models in a particular evaluation direction, with the bolded numbers representing the best results in that evaluation direction.

Figure \ref{figure2} presents the corresponding results of different models in four evaluation directions. Table \ref{table1} shows the specific performance of different models in a particular evaluation direction, with the bolded numbers representing the best results in that evaluation direction. The results reveal that GPT-4-Turbo-04-09 performs best in syntax (90.69\%) and knowledge (80.52\%) evaluations, while Qwen2-7B-Instruct performs well in semantics (72.53\%). However, all models struggle in reasoning, with the highest accuracy being 29.6\%. These findings highlight the strengths and weaknesses of current LLMs in Mongolian, providing insights for future research and development.

\section{Discussion}
Our Dual Capability Framework categorizes LLM abilities into linguistic and cognitive capabilities, divided into syntax, semantics, knowledge, and reasoning levels. Most modern LLMs, whether open- or closed-source, are trained on multilingual corpora, granting them some degree of multilingual competence. Studying their multilingual performance and its sources is essential for advancing LLMs.

This study examines LLM performance in Mongolian across different levels. Our experiments show that while models perform well in basic linguistic tasks, they struggle with semantic understanding and complex reasoning. Aligning knowledge content with the models' main training language could improve their performance in low-resource languages.

Multilingual capability remains a critical research area. Our study suggests that while models possess basic skills, performance varies by language and task. Future work should explore the mechanisms behind multilingual competence and ways to improve it.

MM-Eval evaluates LLMs hierarchically but is limited by its single content source, only multiple-choice questions, and a narrow scope of logical reasoning tasks. Expanding these areas will enable more comprehensive evaluations.

\bibliography{custom}

\begin{thebibliography}{21}
\providecommand{\natexlab}[1]{#1}

\bibitem[{Chen et~al.(2021)Chen, Tworek, Jun, Yuan, de~Oliveira~Pinto et~al.}]{HumanEval}
Mark Chen, Jerry Tworek, Heewoo Jun, Qiming Yuan, Henrique~Pond{\'{e}} de~Oliveira~Pinto, et~al. 2021.
\newblock \href {https://arxiv.org/abs/2107.03374} {Evaluating large language models trained on code}.
\newblock \emph{CoRR}, abs/2107.03374.

\bibitem[{Cheng et~al.(2020)Cheng, Wang, Bao, and Gao}]{named}
Xiao Cheng, Weihua Wang, Feilong Bao, and Guanglai Gao. 2020.
\newblock \href {https://doi.org/10.1007/978-981-33-6162-1\_2} {{MTNER:} {A} corpus for mongolian tourism named entity recognition}.
\newblock In \emph{Machine Translation - 16th China Conference, {CCMT} 2020, Hohhot, China, October 10-12, 2020, Revised Selected Papers}, volume 1328 of \emph{Communications in Computer and Information Science}, pages 11--23. Springer.

\bibitem[{Cobbe et~al.(2021)Cobbe, Kosaraju, Bavarian, Chen, Jun et~al.}]{GSM8K}
Karl Cobbe, Vineet Kosaraju, Mohammad Bavarian, Mark Chen, Heewoo Jun, et~al. 2021.
\newblock \href {https://arxiv.org/abs/2110.14168} {Training verifiers to solve math word problems}.
\newblock \emph{CoRR}, abs/2110.14168.

\bibitem[{Dai et~al.(2024)Dai, Deng, Zhao, Xu, Gao, Chen, Li, Zeng et~al.}]{deepseek}
Damai Dai, Chengqi Deng, Chenggang Zhao, R.~X. Xu, Huazuo Gao, Deli Chen, Jiashi Li, Wangding Zeng, et~al. 2024.
\newblock \href {https://aclanthology.org/2024.acl-long.70} {Deepseekmoe: Towards ultimate expert specialization in mixture-of-experts language models}.
\newblock In \emph{Proceedings of the 62nd Annual Meeting of the Association for Computational Linguistics (Volume 1: Long Papers), {ACL} 2024, Bangkok, Thailand, August 11-16, 2024}, pages 1280--1297. Association for Computational Linguistics.

\bibitem[{Hendrycks et~al.(2021)Hendrycks, Burns, Basart, Zou, Mazeika et~al.}]{MMLU}
Dan Hendrycks, Collin Burns, Steven Basart, Andy Zou, Mantas Mazeika, et~al. 2021.
\newblock \href {https://openreview.net/forum?id=d7KBjmI3GmQ} {Measuring massive multitask language understanding}.
\newblock In \emph{9th International Conference on Learning Representations, {ICLR} 2021, Virtual Event, Austria, May 3-7, 2021}.

\bibitem[{Hou et~al.(2017)Hou, Wang, Yuan, and Liu}]{MMLT1}
Wanzhuang Hou, Hao Wang, Lin Yuan, and Dinan Liu. 2017.
\newblock \emph{Modern Mongolian Language Textbook I}.
\newblock Peking University Press.

\bibitem[{Jin et~al.(2024)Jin, Han, Yang, Jiang, Liu, Chang, Chen, and Hu}]{context}
Hongye Jin, Xiaotian Han, Jingfeng Yang, Zhimeng Jiang, Zirui Liu, Chia{-}Yuan Chang, Huiyuan Chen, and Xia Hu. 2024.
\newblock \href {https://openreview.net/forum?id=nkOMLBIiI7} {{LLM} maybe longlm: Selfextend {LLM} context window without tuning}.
\newblock In \emph{Forty-first International Conference on Machine Learning, {ICML} 2024, Vienna, Austria, July 21-27, 2024}.

\bibitem[{Malik et~al.(2024)Malik, Mayhew, Piech, and Bicknell}]{qg}
Ali Malik, Stephen Mayhew, Christopher Piech, and Klinton Bicknell. 2024.
\newblock \href {https://aclanthology.org/2024.findings-acl.926} {From tarzan to tolkien: Controlling the language proficiency level of llms for content generation}.
\newblock In \emph{Findings of the Association for Computational Linguistics, {ACL} 2024, Bangkok, Thailand and virtual meeting, August 11-16, 2024}, pages 15670--15693. Association for Computational Linguistics.

\bibitem[{OpenAI(2023)}]{gpt4}
OpenAI. 2023.
\newblock \href {https://doi.org/10.48550/ARXIV.2303.08774} {{GPT-4} technical report}.
\newblock \emph{CoRR}, abs/2303.08774.

\bibitem[{Rein et~al.(2023)Rein, Hou, Stickland, Petty, Pang et~al.}]{GPQA}
David Rein, Betty~Li Hou, Asa~Cooper Stickland, Jackson Petty, Richard~Yuanzhe Pang, et~al. 2023.
\newblock \href {https://doi.org/10.48550/ARXIV.2311.12022} {{GPQA:} {A} graduate-level google-proof q{\&}a benchmark}.
\newblock \emph{CoRR}, abs/2311.12022.

\bibitem[{Schimanski et~al.(2024)Schimanski, Ni, Kraus, Ash, and Leippold}]{QA}
Tobias Schimanski, Jingwei Ni, Mathias Kraus, Elliott Ash, and Markus Leippold. 2024.
\newblock \href {https://aclanthology.org/2024.acl-long.105} {Towards faithful and robust {LLM} specialists for evidence-based question-answering}.
\newblock In \emph{Proceedings of the 62nd Annual Meeting of the Association for Computational Linguistics (Volume 1: Long Papers), {ACL} 2024, Bangkok, Thailand, August 11-16, 2024}, pages 1913--1931. Association for Computational Linguistics.

\bibitem[{Shi et~al.(2023)Shi, Suzgun, Freitag, Wang, Srivats et~al.}]{MGSM}
Freda Shi, Mirac Suzgun, Markus Freitag, Xuezhi Wang, Suraj Srivats, et~al. 2023.
\newblock \href {https://openreview.net/forum?id=fR3wGCk-IXp} {Language models are multilingual chain-of-thought reasoners}.
\newblock In \emph{The Eleventh International Conference on Learning Representations, {ICLR} 2023, Kigali, Rwanda, May 1-5, 2023}.

\bibitem[{Song et~al.(2024)Song, Su, Shalyminov, Cai, and Mansour}]{summarization}
Hwanjun Song, Hang Su, Igor Shalyminov, Jason Cai, and Saab Mansour. 2024.
\newblock \href {https://aclanthology.org/2024.acl-long.51} {Finesure: Fine-grained summarization evaluation using llms}.
\newblock In \emph{Proceedings of the 62nd Annual Meeting of the Association for Computational Linguistics (Volume 1: Long Papers), {ACL} 2024, Bangkok, Thailand, August 11-16, 2024}, pages 906--922. Association for Computational Linguistics.

\bibitem[{Touvron et~al.(2023)Touvron, Martin, Stone, Albert, Almahairi, Babaei, Bashlykov et~al.}]{llama2}
Hugo Touvron, Louis Martin, Kevin Stone, Peter Albert, Amjad Almahairi, Yasmine Babaei, Nikolay Bashlykov, et~al. 2023.
\newblock \href {https://doi.org/10.48550/ARXIV.2307.09288} {Llama 2: Open foundation and fine-tuned chat models}.
\newblock \emph{CoRR}, abs/2307.09288.

\bibitem[{White et~al.(2024)White, Dooley, Roberts, Pal, Feuer et~al.}]{LiveBench}
Colin White, Samuel Dooley, Manley Roberts, Arka Pal, Benjamin Feuer, et~al. 2024.
\newblock \href {https://doi.org/10.48550/ARXIV.2406.19314} {Livebench: {A} challenging, contamination-free {LLM} benchmark}.
\newblock \emph{CoRR}, abs/2406.19314.

\bibitem[{Xu et~al.(2023)Xu, Liu, Yan, Xu, Si et~al.}]{CValues}
Guohai Xu, Jiayi Liu, Ming Yan, Haotian Xu, Jinghui Si, et~al. 2023.
\newblock \href {https://doi.org/10.48550/ARXIV.2307.09705} {Cvalues: Measuring the values of chinese large language models from safety to responsibility}.
\newblock \emph{CoRR}, abs/2307.09705.

\bibitem[{Xu et~al.(2024)Xu, Sharaf, Chen, Tan, Shen, Durme, Murray, and Kim}]{icml}
Haoran Xu, Amr Sharaf, Yunmo Chen, Weiting Tan, Lingfeng Shen, Benjamin~Van Durme, Kenton Murray, and Young~Jin Kim. 2024.
\newblock \href {https://openreview.net/forum?id=51iwkioZpn} {Contrastive preference optimization: Pushing the boundaries of {LLM} performance in machine translation}.
\newblock In \emph{Forty-first International Conference on Machine Learning, {ICML} 2024, Vienna, Austria, July 21-27, 2024}.

\bibitem[{Yang et~al.(2024)Yang, Yang, Hui, Zheng, Yu et~al.}]{qwen}
An~Yang, Baosong Yang, Binyuan Hui, Bo~Zheng, Bowen Yu, et~al. 2024.
\newblock \href {https://doi.org/10.48550/ARXIV.2407.10671} {Qwen2 technical report}.
\newblock \emph{CoRR}, abs/2407.10671.

\bibitem[{Yang et~al.(2022)Yang, Xu, Cui, Wang, Lin et~al.}]{cino}
Ziqing Yang, Zihang Xu, Yiming Cui, Baoxin Wang, Min Lin, et~al. 2022.
\newblock \href {https://aclanthology.org/2022.coling-1.346} {{CINO}: A {C}hinese minority pre-trained language model}.
\newblock In \emph{Proceedings of the 29th International Conference on Computational Linguistics}, pages 3937--3949, Gyeongju, Republic of Korea. International Committee on Computational Linguistics.

\bibitem[{Yih et~al.(2016)Yih, Richardson, Meek, Chang, and Suh}]{WebQSP}
Wen{-}tau Yih, Matthew Richardson, Christopher Meek, Ming{-}Wei Chang, and Jina Suh. 2016.
\newblock \href {https://doi.org/10.18653/V1/P16-2033} {The value of semantic parse labeling for knowledge base question answering}.
\newblock In \emph{Proceedings of the 54th Annual Meeting of the Association for Computational Linguistics, {ACL} 2016, August 7-12, 2016, Berlin, Germany, Volume 2: Short Papers}. The Association for Computer Linguistics.

\bibitem[{Zeng et~al.(2023)Zeng, Liu, Du, Wang, Lai et~al.}]{glm}
Aohan Zeng, Xiao Liu, Zhengxiao Du, Zihan Wang, Hanyu Lai, et~al. 2023.
\newblock \href {https://openreview.net/forum?id=-Aw0rrrPUF} {{GLM-130B:} an open bilingual pre-trained model}.
\newblock In \emph{The Eleventh International Conference on Learning Representations, {ICLR} 2023, Kigali, Rwanda, May 1-5, 2023}.

\end{thebibliography}

\appendix

\section{MM-Eval Dataset Examples}
\label{sec:appendix}

\subsection{Syntax Eval}
\begin{figure}[H]
    \centering
     \caption{Syntax Eval Examples}
    \includegraphics[width=\linewidth]{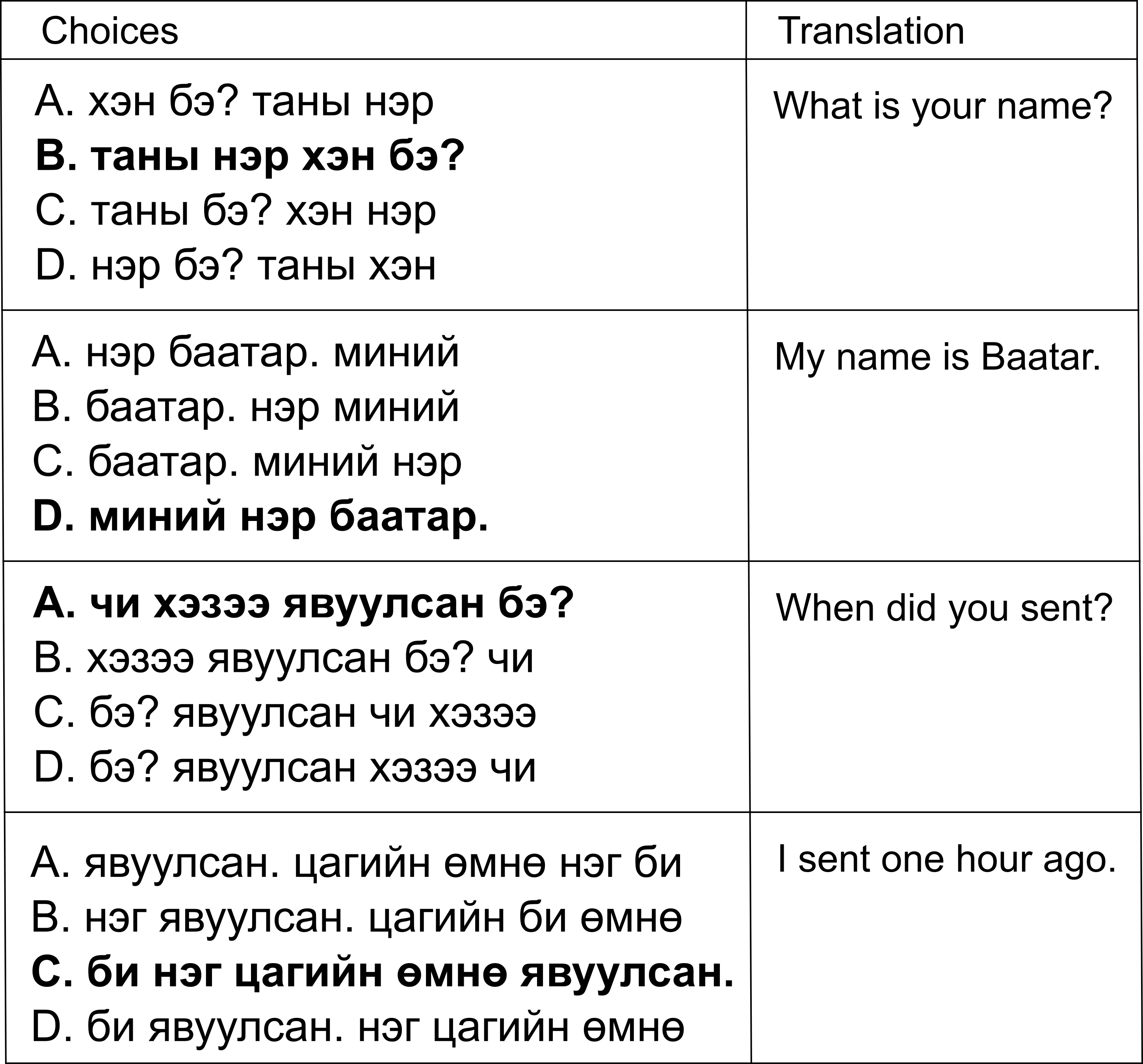}
\end{figure}

\subsection{Semantics Eval}
\begin{figure}[H]
    \centering
    \caption{Semantics Eval Examples}
    \includegraphics[width=\linewidth]{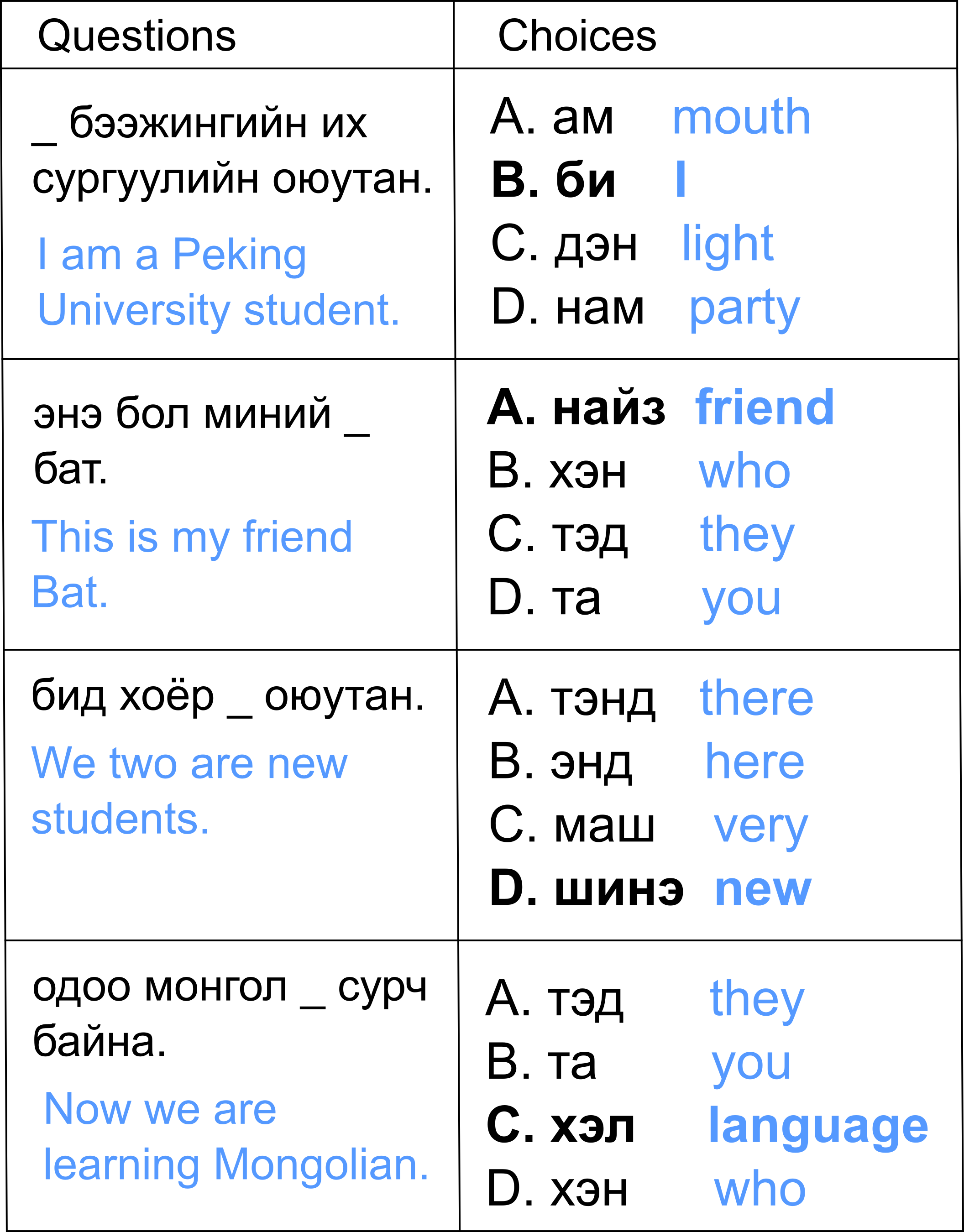}
\end{figure}

\subsection{Knowledge Eval}
\begin{figure}[H]
    \centering
    \caption{Knowledge Eval Examples}
    \includegraphics[width=\linewidth]{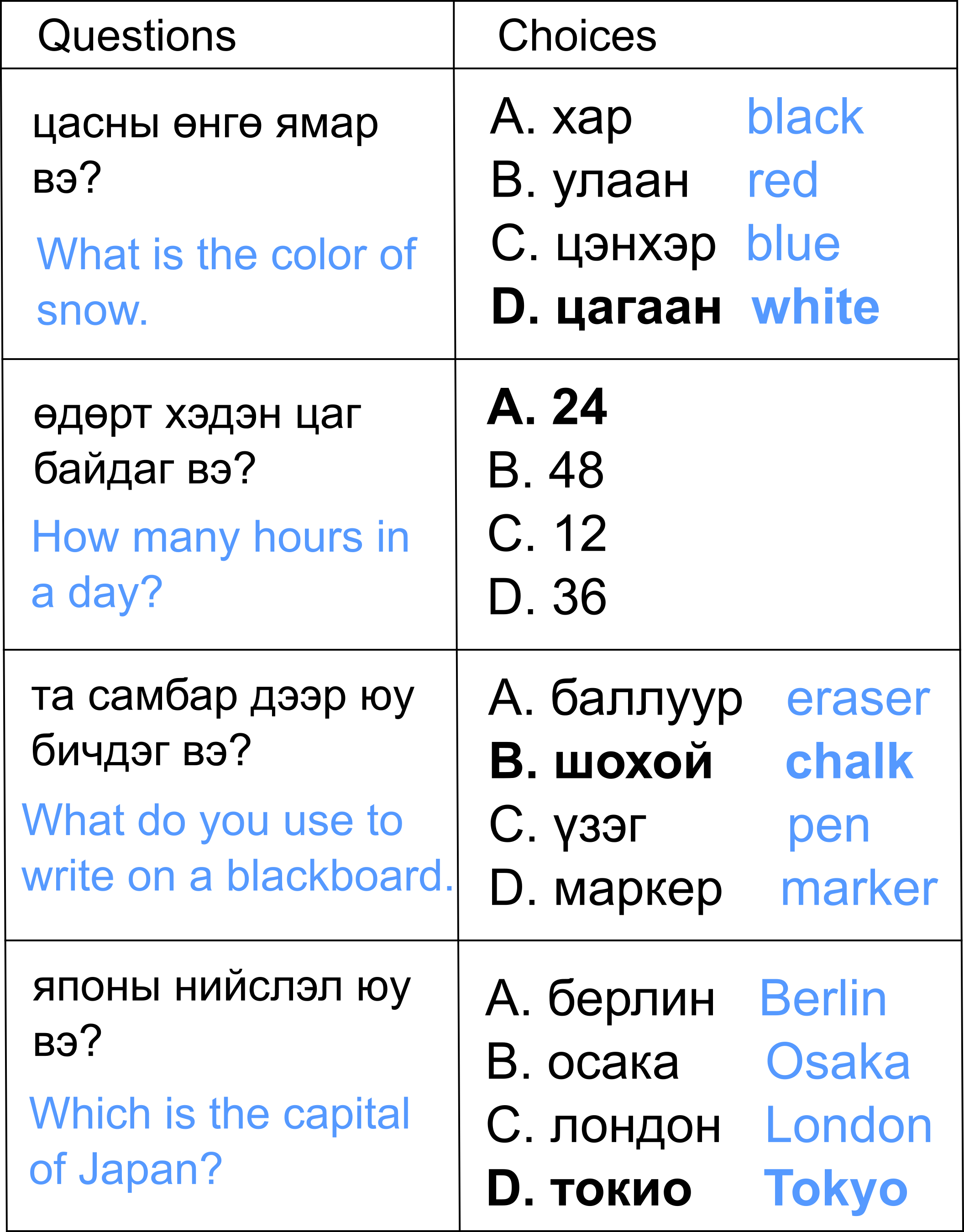}
\end{figure}

\subsection{Reasoning Eval}
\begin{figure}[H]
    \centering
     \caption{Reasoning Eval Examples}
    \includegraphics[width=\linewidth]{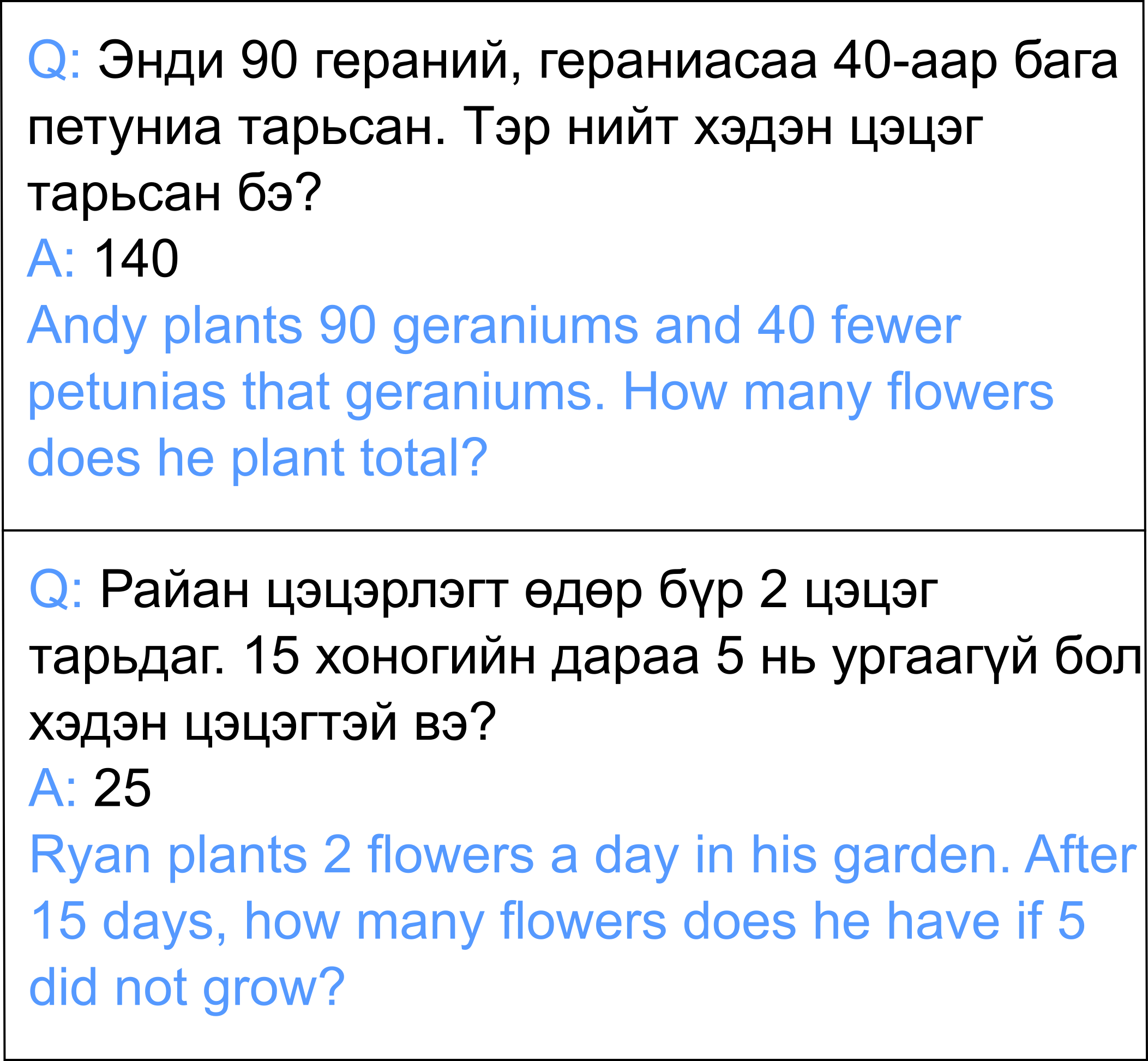}
\end{figure}

\end{document}